\documentclass[a4paper,conference]{IEEEtran}
\usepackage{cite}

\usepackage{graphicx}
\usepackage{amsmath}
\usepackage{mathtools}
\usepackage{amssymb}
\usepackage{algorithmic}

\usepackage{array}
\usepackage{fixltx2e}
\usepackage{dblfloatfix}

\usepackage{url}
\usepackage{float}

\usepackage{multirow}

\usepackage{xcolor}
\newcommand{\etal}{\emph{et al. }}

\begin{document}
%
\title{UDBNET: Unsupervised Document Binarization Network \textit{via} Adversarial Game}


\author{Amandeep Kumar\textsuperscript{1*} \hspace{.2cm} Shuvozit Ghose\textsuperscript{2*} \hspace{.2cm} Pinaki Nath Chowdhury\textsuperscript{3} \hspace{.2cm} Partha Pratim Roy\textsuperscript{4}\hspace{.2cm}   Umapada Pal\textsuperscript{5}\hspace{.2cm} \\
\textsuperscript{1}Techno Main Salt Lake, Sector V, Kolkata, India.\hspace{.08cm} \textsuperscript{2}Institute of Engineering and Management Kolkata, India .\\
\textsuperscript{4}Indian Institute of Technology Roorkee, India .
\textsuperscript{3,5}Indian Statistical Institute Kolkata, India .\\
{\tt\small kumar.amandeep015@gmail.com\textsuperscript{1}, shuvozit.ghose@gmail.com\textsuperscript{2}}
}


%


\maketitle

\begin{abstract}
$\let\thefootnote\relax\footnote{* Authors contributed equally}$
Degraded document image  binarization is one of the most challenging tasks in the domain of document image analysis. In this paper, we present a novel approach towards document image binarization by introducing three-player min-max adversarial game. We train the network in an unsupervised setup by assuming that we do not have any paired-training data. In our approach, an Adversarial Texture Augmentation Network (ATANet) first superimposes the texture of a degraded reference image over a clean image. Later, the clean image along with its generated degraded version constitute the pseudo paired-data which is used to train the Unsupervised Document Binarization Network (UDBNet). Following this approach, we have enlarged the document binarization datasets as it generates multiple images having same content feature but different textual feature. These generated noisy images are then fed into the UDBNet to get back the clean version. The joint discriminator which is the third-player of our three-player min-max adversarial game tries to couple both the ATANet and UDBNet. The three-player min-max adversarial game stops, when the distributions modelled by the ATANet and the UDBNet align to the same joint distribution over time. Thus, the joint discriminator enforces the UDBNet to perform better on real degraded image. The experimental results indicate the superior performance of the proposed model over existing state-of-the-art algorithm on widely used DIBCO datasets. The source code of the proposed system is publicly available at https://github.com/VIROBO-15/UDBNET.
\end{abstract}


%
\IEEEpeerreviewmaketitle

\section{Introduction}
 Document image binarization is a rudimentary problem in the field of Document analysis. Binarization itself, is the prepossessing backbone of many document image processing systems (DIPSs)\cite{chen2017broken,jia2016document}. The performance of the high level processing tasks, such as image segmentation \cite{otsu1979threshold}, line segmentation \cite{roy08}, word recognition \cite{Chherawala16, BhuniaCVPR}, optical character recognition (OCR) \cite{Keserwani19}, and document layout analysis (DLA) \cite{tran2017robust} is greatly dependent on the success of the binarization task. Technically, document image binarization is the technique of converting color document images or gray-level images into a binary representation, where the main objective is to classify each pixel as foreground(text/ink) or background(parchment/paper). In other words, it is the process of discarding the unnecessary noisy information while preserving the meaningful visual information.
 
 Document image binarization can be considered as an easy task for images of uniform distribution. However, in real-world scenarios under significant image noise and uneven background, binarization is a quite challenging problem. Moreover, the document images suffer from various degradation due to faint characters, bleed-through background, clutter and artifacts, dark patches, creases, faded ink, non-uniform variation of intensity, inadequate maintenance, aging effect, ink stains, lighting conditions, warping effect during acquisition etc. \cite{Roy15, Konwer18}. Faded ink creates difficulty during distinguishing light text from background. 
 Bleed through occurs when content from the back of a page becomes visible or `leaks' through. It creates difficulty in labeling foreground and background during binarization process as it can misinterpret background as foreground. Uneven illumination happens when the image is suffering from shadow effect or inconsistent lighting during acquisition. In addition to the above, dark patches are quite difficult to remove for various reasons. Firstly, these patches are of varying sizes and intensities. Secondly, they appear as stains of arbitrary shapes. Thirdly, they are often present in areas containing characters. Therefore, the study on binarization for document images, specially in the context of degraded images, is highly essential. 
 
 In general, binarization methods \cite{vo2018binarization} \cite{westphal2018document} \cite{he2019deepotsu} works for supervised setup. In the supervised setup, we need ground-truth binarized image along with the degraded image. But, it is difficult to get the corresponding ground truth binary image in many scenarios like in case of historical document image. To address these drawbacks, Bhunia \etal \cite{bhunia2019improving} first attempts to introduce unsupervised setup in the domain of document image binarization. For this purpose, they employ Texture Augmentation Network (TANet) that superimposes the noisy appearance of the degraded document on the clean binary image to generate multiple degraded image of same textual content with various noisy textures and later utilize Binarization Network (BiNet) to get back the clean version of the document image. Although this method has shown better results over the previous state-of-the-art methods, it has several limitations. Firstly, the TANet is completely unaware about the content at which it is conditioned on. Thus, the corresponding discriminator can not verify if the content of the generated noisy image remain consistent or not. Secondly, there exist no performance quantifier that validates the performance of the BiNet on real degraded noisy image. Finally, the Binarization Network (BiNet) has dataset bias towards generated noisy images. But, to adddress the dataset bias, BiNet does not use any kind of formulation or other techniques. In our observation, these limitations are due to the fact that the TANet and BiNet both employ straight-forward two-player Generative Adversarial Network (GAN) \cite{goodfellow2014generative} objectives and model two different uncorrelated conditional distributions. In this paper, we address these limitations by introducing adversarial min-max game in the domain of unsupervised document image binarization. Similar to the TANet and BiNet, we propose Adversarial Texture Augmentation Network (ATANet) and Unsupervised Documenet Binarization Network (UDBNet) which utilize three-player GAN objectives. The proposed third player is a joint discriminator tries to couple both the Adversarial Texture Augmentation Network (ATANet) and Unsupervised Document Binarization Network (UDBNet). Our three-player min-max adversarial game comes to an end, when the distribution modelled by the Adversarial Texture Augmentation Network (ATANet) and the Unsupervised Document Binarization Network (UDBNet) align to the same joint distribution over time.
 Therefore, the contributions of this paper are as follows:
 
 \begin{itemize}
 \item To be our best of knowledge, we are the first one to introduce adversarial game in the domain of document image binarization by proposing Adversarial Texture Augmentation Network (ATANet) and Unsupervised Document Binarization Network (UDBNet).

 \item We introduce a joint discriminator which tries to couple the ATANet and UDBNet so that it can tackle the dataset bias problem and perform well on the real degraded document image.

 \item Our approach shows a superior performance on widely used DIBCO datasets as compared to the existing state-of-the-arts methods.
 \end{itemize}
 The remaining of the paper is organized as follows: In section \ref{related}, we discuss about the related works in the field of document image binarization. In section \ref{section:proposed_method},  we describe the proposed framework. The datasets, implementation, baselines methods and performance analysis are discussed in section \ref{experiment}. Section \ref{conclusion} concludes the paper.

\section{Related Works} \label{related}
Document image binarization is a classical research problem in computer-aided document analysis and has been studied extensively over the past few decades. Document image binarization aims at converting the document image into either foreground text or background. The most simple and widely used approach is thresholding, which sets the pixels under a threshold value to 0 and the rest to 1.  Thresholding methods are primarily of three types : global, local and hybrid. In high quality images, global algorithms can effectively estimate a threshold based on the entire image.  Global thresholds can be calculated using gray level histogram \cite{otsu1979threshold},  circular statistics \cite{lai2014efficient}, error minimization \cite{kittler1986minimum}, histogram entropy\cite{kapur1985new} and  moment preserving principle\cite{tsai1985moment}. Clustering models\cite{papamarkos2001technique} can also learn mappings in an unsupervised manner based on global features to separate background and foreground. However performance degrades when they are applied to images having variations in background due to illumination, occlusion or degradation. For such cases, local adaptive methods perform better. Some of the common local thresholding approaches can be seen in the works of Bernsen \etal \cite{bernsen1986dynamic}, Niblack \etal \cite{niblack1985introduction}, Sauvota \etal \cite{sauvola2000adaptive}.   
 In the work by Niblack \etal \cite{niblack1985introduction}, a major drawback is that if the foreground text is sparse, a lot of background noise will remain in the binary image. Sauvota \etal \cite{sauvola2000adaptive} alleviates this by assuming the foreground pixels to be closer to background ones.

Apart from threshold based techniques, non-threshold based strategies have also been studied extensively in literature. Some notable approaches include Markov Random Field (MRF) modeling of an input image, which minimizes a cost function by regarding the target binarized image as a binary MRF. Howe \etal \cite{howe2011laplacian} proposed an algorithm where they defined the cost function based on combination of the Laplacian energy of image intensity for computing local likelihood of foreground-background pixels and Canny edge detection for detecting discontinuities. The cost function is minimized by graph cut computation. Howe’s method\cite{howe2011laplacian} is efficient and yields good results, however it is parameter dependent. Howe \etal \cite{howe2013document} improvised on this method by adaptive tuning of two parameters to yield better performance. Howe’s technique formed the basis of the first winning algorithm proposed by Kliger and Tal in the DIBCO 2016 competition\cite{pratikakis2016icfhr2016}. They combined Howe’s algorithm with a novel pre-processing step based on linear transformation of the image onto a spherical surface where concavities correspond to foreground in the original image.The concavities are estimated using the Hidden Point Removal Operator\cite{katz2007direct} which outputs a probability of a pixel belonging to a concavity. 

All these proposed techniques perform well in the context they are applied to. But these methods fail to generalize in the context of binarizing any kind of document subjected to a varied degree of illumination, background noise and degradation.
Recently pixel-wise binarization approaches have been proposed in literature where each pixel is classified as text or background. Pastor-Pellicer\cite{pastor2015insights} proposed a CNN framework consisting of two groups of convolution layers and a fully connected layer. Each pixel is classified into text or background by using a sliding window centred at the classified pixel. Such an approach has also been used in binarizing musical documents by Calvo-Zaragoza \etal \cite{calvo2017pixel} . These pixel wise classification techniques have shown good performance, however their most conspicuous drawbacks include being computationally very expensive since they involve labelling each pixel in the document image and classifying each pixel independently without exploiting contextual information in any pixel’s neighbourhood. To incorporate this contextual information, Afzal \etal \cite{afzal2015document} propose a pixel wise classification method where they formulate the binarization procedure as a sequence learning problem. They use a 2D LSTM model which takes in a 2D sequence of pixels as input and classifies each pixel as foreground or background. This achieved better results but still suffered from huge computational complexity. To alleviate this, Tensmeyer \etal \cite{tensmeyer2017document} proposed a novel multi-scale fully convolutional network for document image binarization. Recently Calvo-Zaragoz \etal \cite{calvo2019selectional} proposed a fully convolutional-selectional auto encoder model that has been trained to learn a patch-wise mapping of the document image to its corresponding binarized version. This performs a fine-grained categorization in which each pixel gets a different activation value depending on whether the target label of the pixel is text or background. Other approaches involving convolution networks include the winning algorithm of DIBCO 2017 competition\cite{pratikakis2017icdar2017}, where the winning team used a U-Net encoder decoder architecture for accurate pixel classification.  
Vo \etal \cite{vo2018binarization} introduced a hierarchical deep supervised network for document binarization which achieves state of the art performance on several benchmark datatsets. Westphal \etal \cite{westphal2018document} proposed a Grid LSTM network for binarization, yet it achieves lesser performance than Vo’s method\cite{vo2018binarization}. To learn the document degradation, He \etal \cite{he2019deepotsu} proposed an iterative fine tuning technique to learn the mappings from a degraded input document image to the expected clean and uniform images followed by a classifier to output the binarized image. 

In case of unsupervised image-to-image translation task, one of the first major works that uses deep network is CycleGAN \cite{zhu2017unpaired}. Following this work, there have been numerous attempts to design unsupervised or semi-supervised framework for different computer vision tasks like depth estimation \cite{zheng2018t2net}, image captioning \cite{gu2019unpaired, feng2019unsupervised} etc. Most of these works use popular cycle-consistency loss to learn an unsupervised mapping between two different domains. In contrast to all such works, Bhunia \etal \cite{bhunia2019improving} employ Texture Augmentation Network (TANet) that superimposes the noisy appearance of the degraded document on the clean binary image to generate multiple degraded image of same textual content with various noisy textures and later utilize Binarization Network (BiNet) to get back the clean version of the document image.

\begin{figure*}[]
	\begin{center}
		\includegraphics[width=1\linewidth]{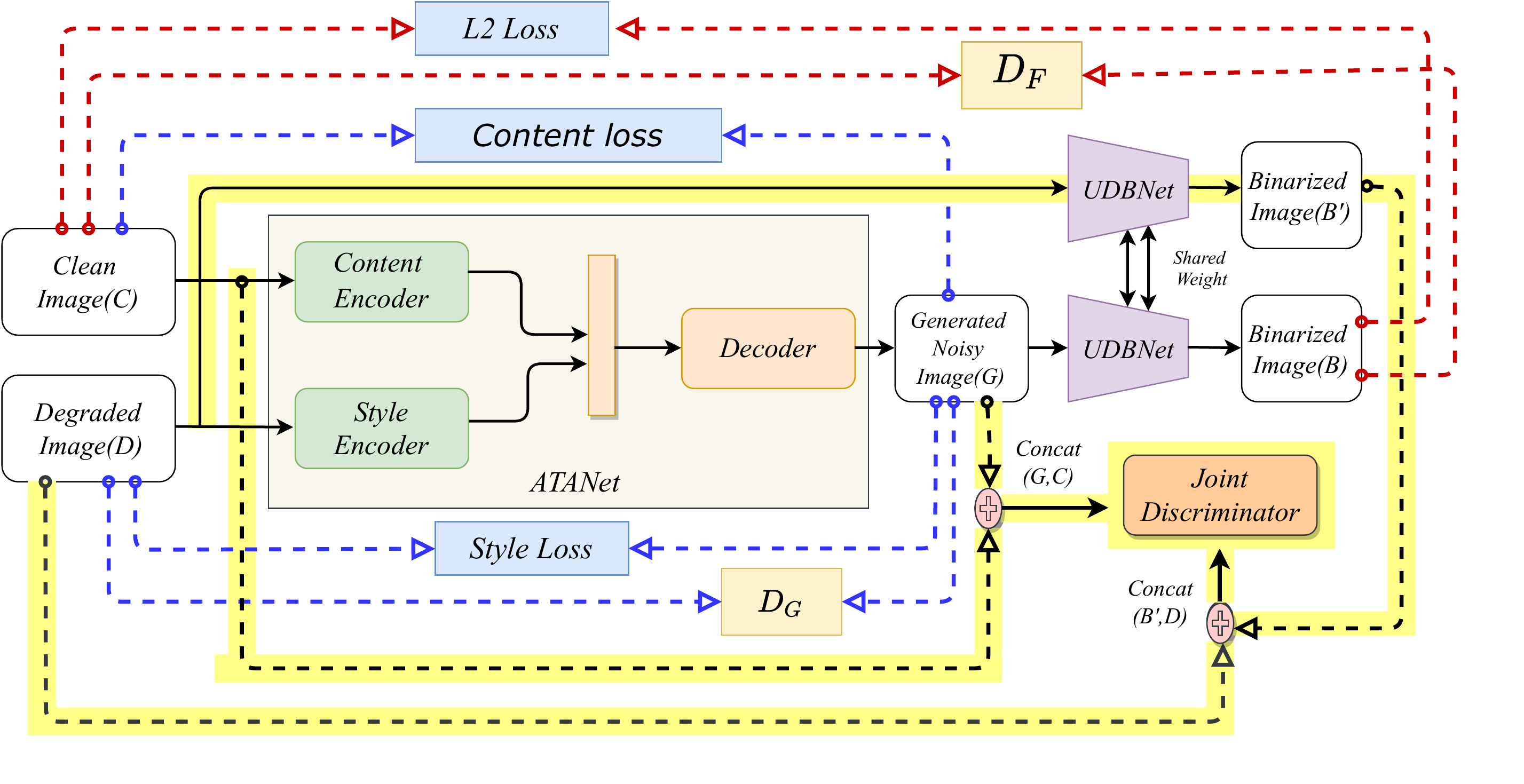}
	\end{center}
	\vspace{-0.35in}
	\caption{Illustration of Our proposed Framework. The yellow highlighted region highlights our contribution over Bhunia \etal \cite{bhunia2019improving}. We have two networks: ATANet which takes Clean image $C$ and Degraded image $D$ as inputs and generates noisy degraded image $G$. On the other hand, UDBNet tries to get back the clean image from the generated noisy $G$. Thus, $(C,G)$ acts as pseudo paired data to train the binarization network. Also,  We feed degraded image $D$ as an input to UDBNet and get the corresponding binarized image$(B')$. Then, we concat the image pairs $(G,C)$ and $(B',D)$ and feed into joint discriminator to couple both ATANet and UDBNet. This enforces the UDBNet to generalize better for real degraded images}
	\label{fig:img2}
	\vspace{-.05in}
\end{figure*}

\section{Proposed Framework}\label{section:proposed_method}
In this section. we first briefly present the binarization model proposed by Bhunia \etal \cite{bhunia2019improving} as base model. Next, we describe the limitations of the base model. Finally, we introduce our novel unsupervised adversarial game and describe how Adversarial Texture Augmentation Network (ATANet) and Unsupervised Document Binarization Network (UDBNet) address these limitations efficiently.
 
\subsection{Background: Base Models}
The base model consists of two networks: Texture Augmentation Network(TANet) and Binarization Network(BiNet). Let $C$ denotes the binarized clean image sampled from marginal distribution $P(C)$ and $D$ denotes a degraded document image sampled from marginal distribution $P(D)$. TANet tries to model $P(D|C)$, i.e., given a clean image, it tries to generate a degraded version of it keeping the content same. On the other hand, BiNet tries to model $P(C|D)$, i.e., given a degraded image, it tries to generate the clean image. During inference, only BiNet is used. 

\textbf{Texture Augmentation Network} : The TANet exploit a two-player GAN which consists of a generator network and a discriminator network. Conditional distribution $P(D|C)$ is approximately modelled by $P(G|C,D) \approx P(D|C)$, where the noisy texture of degraded image $D$ is superimposed on the clean image $C$ to generate noisy version of the clean image as $G$. Note that, the content of  $G$ and $C$ remains similar and we do not use any paired-data here. On the other side, the discriminator tries to discriminate between output image $G$ and degraded reference image $D$. The generator of the TANet use a content encoder and a style encoder to encode the semantic content of the $C$ image and noisy texture of the $D$ image explicitly. Next, the two encoded features are concatenated to obtain a mixed feature representation. Finally, this mixed representation is passed through a decoder network that outputs noisy generated image $G$. To ensure that the generated image $G$ contains the same textual content as clean image $C$ and the same texture element of the degraded image $D$, The TANet utilizes the following loss functions:

\textit{Adversarial loss:} The objective of the adversarial loss is to constrain the output to make it similar to the degraded reference image $D$. The adversarial loss is defined as:
\begin{equation} 
\begin{split}
    \mathcal{L}_{T}^{GAN}(T, D_T) = \mathbb{E}_{{D}\sim{P(D)}}[\log {D_T}({D})] + \\
\mathbb{E}_{C\sim{P(C)},{D}\sim{P(D)}}[\log (1-{D_T}(T(C,D))]
\end{split}
\end{equation}
Where, the discriminator $D_T$ tries to discriminate between the output image $G$ from the degraded reference image $D$.

\textit{Style loss:} While adversarial loss focuses on getting the overall structure of the generated image, an additional style loss $\mathcal{L}^s(T)$ ensures successful transfer of texture content from degraded reference image $D$ to the input binarized clean image $C$. For this purpose, Gram matrices \cite{gatys2015texture, gatys2016image} is used in \textit{``conv1\_1", ``conv2\_1",``conv3\_1",``conv4\_1",``conv5\_1"} layers of the encoder networks. Mathematically, 
\begin{equation}
\begin{split}
\mathcal{G}^{l}_{ij} = \sum_{k}{\mathcal{F}^{l}_{ik}{\mathcal{F}^{l}_{jk}}}
\end{split}
\end{equation}
Where, $F^{l}_{ik}$ is the activation of $i^{th}$ filter at position k in layer l, gram matrix $\mathcal{G}^{l}_{ij} \in \mathbb{R}^{N_l \times N_l}$ is the inner product between vectorised feature maps $i$ and $j$ in layer $l$ and $N_l$ is the number of feature maps.

\textit{Content loss:} To ensure the generated image $G$ contains the same textual content as the clean binarized image $C$, a content loss is defined as follows:
\begin{equation}
\begin{split}
\mathcal{L}^c(T) = ||M\odot C - M\odot G||_2
\end{split}
\end{equation}
Here, $M$ denotes a binary mask that has value 0 in the background and 1 in the text region.

The overall objective of the TANet is defined as follows:
\begin{equation}
\begin{split}
\mathcal{L}^{TANet} = \mathcal{L}^{GAN}_{T}(T, D_T) + \lambda_s \mathcal{L}^s(T) + \lambda_c \mathcal{L}^c(T)
\end{split}
\end{equation}
Where, $\lambda_s$ and $\lambda_c$ are the tunable hyper-parameters to balance multiple objectives.

\textbf{Binarization Network} : Similar to TANet, BiNet exploits two-player GAN and employs an image-to-image translation framework consisting of a generator and a discriminator. While the generator of the BiNet tries to model $P(B|G)\approx P(C|D)$, where $B$ is the binarized clean image of the newly generated noisy image, the discriminator determines how good the generator is in generating binarized images. The adversarial loss of the BiNet is:

\begin{equation} 
\begin{split}
\mathcal{L}_{F}^{GAN}(F,D_F) = \mathbb{E}_{C \sim P(C)}[log D_F(C)] + \\
\mathbb{E}_{G \sim P(D|C)}[\log(1-D_F(F(G)))]
\end{split}
\end{equation}

In times of training, for each input image $G$, there is corresponding ground truth image $C$. Thus, an additional $L_2$ loss is utilized to fully supervised the predicted binarization results along with the adversarial loss:
\begin{equation}
\begin{split}
\mathcal{L}^{L2} = || C - B||_2
\end{split}
\end{equation}
While the $L_2$ pixel loss helps to preserve the content, the adversarial loss guides to obtain sharper output image $B$ by de-noising input noisy image $G$.

The overall objective of the BiNet is as follows:
\begin{equation}
\begin{split}
\mathcal{L}^{BINet} = \mathcal{L}^{GAN}_{F}(F, D_F) + \lambda_{L2}\mathcal{L}^{L2}(F) 
\end{split}
\end{equation}
Where, $\lambda_{L2}$ is a tunable hyper-parameter. Although we show the architecture of our base model in Figure \ref{fig:img2}, We refer the reader to the original work \cite{bhunia2019improving} for further details.

\subsection{Limitations of Base Model}\label{section:Limitations of Base Model}
The limitations of the base model are given below:

\textbf{Limitation 1.} Although the texture augmentation network (TANet) tries to model $P(D|C)$ in the base model and generates noisy images, but it is completely unaware about the content at which it is conditioned on. Thus, the corresponding discriminator can not verify if the content of the generated noisy image remains consistent or not.  

\textbf{Limitation 2.} For unpaired real degraded noisy image, the corresponding ground truth or binarized clean image is absent. The absence of ground truth image limits the scope of binarization network (BiNet) of the base model. Firstly, the BiNet can not be trained with real degraded noisy image as $L_2$ loss can not be utilized. Secondly, since the TANet and the BiNet model two different uncorrelated conditional distribution and are trained separately, these models are prone to overfitting. Finally, there exists no performance quantifier that validates the performance of the BiNet on real degraded noisy image.

\textbf{Limitation 3.} There exist a gap between generated noisy image distribution and real degraded noisy image distribution. As the Binarization network (BiNet) is completely trained on generated noisy image, the Binet has dataset bias towards generated noisy images. A model trained in the generated data can hardly perform well on the real data. This problem is quite similar to domain-shift \cite{quinonero2008covariate} problem. But to minimize the generated-real domain shift in the context of document image binarization, the base model does not use any kind of formulation or other techniques.

\subsection{Adversarial Game}

In our unsupervised setup, we do not have any paired training data. Thus, we do not have any access to real joint distribution of clean and degraded image, $P_{real}(C,D)$. However, we can approximate this real distribution by texture augmentation network and binarization network. 
The joint distribution $P(C,D)$ can be factorized in two ways, namely
\begin{equation}
\begin{split}
P(C, D)=\underbrace{P(D|C)*P(C)}_{P_{T}}= \underbrace{P(C|D)*P(D)}_{P_{B}}
\end{split}
\end{equation}

Please note that, $P(D|C)$ is modelled by texture augmentation network and $P(C|D)$ is modelled by binarization network.  $P_T(C,D)$ is obtained when the Generated noisy image $G$ from Adversarial Texture Augmentation network is concatenated with the corresponding input clean image $C$. On the other side, when we pass a real degraded image $D$ through binarization network to get a corresponding clean image $B'$, we constitute $P_B(C,D)$ by concatenating them together. Thus, texture augmentation network plays role in modeling $P_T(C,D)$ and binarization network plays role in modeling $P_B(C,D)$. $P_T(C,D)$ and $P_B(C,D)$ are both approximated joint-distribution of real and degraded images. If we can properly align these two approximated joint distribution $P_B(C,D)$ and  $P_T(C,D)$ together, it will closely get aligned with the real joint-distribution. In order to align $P_B(C,D)$ and  $P_T(C,D)$, we propose joint discriminator that distinguishes whether a input sample is from the distribution $P_T(C,D)$ or the $P_B(C,D)$. The Adversarial Texture Augmentation Network(ATANet) and Unsupervised Document Binarization(UDBNet) network objective is to fool the discriminator such that it cannot distinguishes whether the input sample is from $P_T(C,D)$ or $P_B(C,D)$. Thus, the distributions of $P_T$ and $P_B$ gets aligned overtime.

\textbf{Adversarial Texture Augmentation Network:} 
 The ATANet Consists of three components: 1) a generator $\mathbf{T}$
 that characterizes the conditional distribution $P_T(G|C,D)$ and generates noisy image $G$; 2) a discriminator $\mathbf{D_T}$ that discriminates the output image $G$ from the degraded reference image $D$; 3) a joint discriminator $\mathbf{J_D}$ that distinguishes whether a pair of data $(G,C)$ comes from  $P_T(C,D)$ or $P_B(C,D)$.

Similar to our base model, the generator consists of content encoder and the style encoder in which we pass the clean image $C$ and the degraded reference image $D$ as the input, respectively. The latent representations after the encoding the images are simply concatenated and feed into the decoder. The architecture of decoder is symmetrical to the encoder, having the skip connection between the layers of content encoder and decoder as similar to our base model. The discriminator $\mathbf{D_T}$ tries to discriminate between the generated image $G$ from the generator and the degraded reference image $D$. The pseudo generated-clean image pair $(G,C)$ are fed into the joint discriminator $\mathbf{J_D}$ such that our joint discriminator tries to distinguish whether the input sample $(G,C)$ is from distribution $P_T(C,D)$  or $P_B(C,D)$.

In this game, let a clean-degraded image pair $(C,D)$ is sampled from distributions $P(C)$and $P(D)$, generator $\mathbf{T}$ produces a pseudo generated noisy image $G$ given $C$ following the conditional distribution $P_T(G|C)$. Hence, the pseudo clean-generated image pair is a sample from the joint distribution $P(C,D)=P(C)P_{T}(G|C)$.

To get the real world noisy, degraded document image having the textual appearance similar to degraded reference image $D$ and textual content similar to the clean image $C$.
we define adversarial loss of our ATANet as:

\begin{equation}
\begin{split}
    \mathop{min}_{\mathbf{D_T}} \mathop{max}_{\mathbf{T},\mathbf{J_D}}\mathcal{L}_{T}^{Adv}(\mathbf{D_T}, \mathbf{T} , 
    \mathbf{J_D}) =  \mathbb{E}_{D \sim P_D}[\log {\mathbf{D_T}}({D})] +  \\
    \mathbb{E}_{(C) \sim P(C),(D) \sim P(D)}[\log (1-{\mathbf{D_T}}(\mathbf{T}(C,D))] + \\
    \mathbb{E}_{(C) \sim P(C),(D) \sim P(D)}[\log (1-{\mathbf{J_D}}(\mathbf{T}(C,D),C)] + \\
    \mathbb{E}_{(D) \sim P(D)}[\log ({\mathbf{J_D}}(\mathbf{F}(D),D)]
\end{split}
\end{equation}

Where, $F$ is Unsupervised Document Binarization Network. The adversarial loss is trained in an alternative manner. The game will end when distribution $P_T(C,D)$ and distribution $P_B(C,D)$  will be in equilibrium and gets aligned over time.

Therefore, the overall objective of the ATANet is defined as:
\begin{equation}
\begin{split}
\mathcal{L}^{ATANet} = \mathcal{L}^{Adv}_{T}(D_T, T, J_T) + \lambda_s \mathcal{L}^s(T) + \lambda_c \mathcal{L}^c(T)
\end{split}
\end{equation}
Where,  $\mathcal{L}^s(T)$ and $\mathcal{L}^c(T)$ are style loss and content loss similar to our base mode, $\lambda_s$ and $\lambda_c$ are the tunable hyper-parameters to balance multiple objectives.

\textbf{Unsupervised Document Binarization Network:} Similiar to ATANet, UDBNet consists of three components: 1) a generator $\mathbf{F}$ that characterizes the conditional distribution $P_B(B|G)$ and $P_B(B'|D)$ generates binarized clean image $B$ and $B'$ corresponding to $G$ and $D$ respectively; 2) a discriminator $\mathbf{D_F}$ determines how good the generator is in generating binarized images $B$; 3) a joint discriminator $\mathbf{J_D}$ that distinguishes whether a pair of data $(B',D)$ comes from distribution $P_B(C,D)$ or $P_T(C,D)$.

We have used similar network architecture for the generator and the discriminator as in the base model. The generated noisy image from the ATANet $G$ is fed into the generator of the UDBNet and generates the binarized image $B$. The discriminator tries to discriminate between the binarized image $B$ and the original clean image $C$. The pseudo clean-degraded image pair $(B',D)$ are fed into the joint discriminator $\mathbf{J_D}$ such that our joint discriminator tries to distinguish whether the input sample $(B',D)$ is from distribution $P_T(C,D)$  or $P_B(C,D)$.

In this game, let a clean-degraded image pair $(B',D)$ is sampled from distributions $P(C|D)$ and $P(D)$, generator $\mathbf{F}$ produces a pseudo binarized clean image $B'$ given $D$ following the conditional distribution $P_B(B'|D)$. Hence, the pseudo degraded-clean image pair is a sample from the joint distribution $P(C,D)=P(D)P_B(B'|D)$.

To attain the proper binarized image, We define adversarial loss of our UDBNet as:
\begin{equation}
\begin{split}
    \mathop{min}_{\mathbf{D_F}} \mathop{max}_{\mathbf{F},\mathbf{J_D}}\mathcal{L}_{F}^{Adv}(\mathbf{D_F}, \mathbf{F} , \mathbf{J_F}) =  \mathbb{E}_{C \sim P_C}[\log {\mathbf{D_F}}({C})] + \\
    \mathbb{E}_{G \sim P(D|C)}[\log (1-{\mathbf{D_F}}(\mathbf{F}(G))] + \\
    \mathbb{E}_{(D) \sim P(D)}[\log (1-{\mathbf{J_D}}(\mathbf{F}(D),D)] + \\
    \mathbb{E}_{(C) \sim P(C),(D) \sim P(D)}[\log ({\mathbf{J_D}}(\mathbf{T}(C,D),C)]
\end{split}
\end{equation}
Where, $T$ is an adversarial texture augmentation network. The adversarial loss is trained in an alternative manner. 
Therefore, the overall objective of the UDBNet is defined as:
\begin{equation}
\begin{split}
\mathcal{L}^{UDBNet} = \mathcal{L}^{Adv}_{T}(D_T, T, J_T) + \lambda_{L2}\mathcal{L}^{L2}(F) 
\end{split}
\end{equation}
Where, $\mathcal{L}^{L2}(F)$ is $L_2$ loss,  $\lambda_{L2}$ is a tunable hyper-parameter.

\subsection{Training Joint Discriminator via Flipped Label}
Similar to the Texture Augmentation network (TANet) of the base model, we feed our clean image $C$ and degraded image $D$ into the content encoder and style encoder, respectively. The encoded representations are simply concatenated and passed through the decoder which outputs the noisy version of clean image $G$. The degraded reference image $D$ is then fed into the generator of Unsupervised Document Binarization Network (UDBNet), which generates binarized version of the image $B'$. To feed into the joint discriminator $\mathbf{J_D}$, we perform simple concatenation of the input pairs $(G,C)$ and $(B',D)$. Both the Adversarial Texture Augmentation Network (ATANet) and Unsupervised Document Binarization Network (UDBNet) tries to fool the joint discriminator such that it cannot discriminate whether the input sample is from distribution $P_T(C,D)$ or the distribution $P_B(C,D)$. The joint discriminator $\mathbf{J_D}$ that distinguishes whether a pair of data $(G,C)$  and $(B',D)$ come from distribution $P_T(C,D)$ or $P_B(C,D)$. This enforces the UDBNet to generalize better for real degraded images although we are not using any paired-training data. The joint discriminator is trained using the flipped labels as utilized in \cite{tzeng2017adversarial}.

\section{Experiment}\label{experiment}

\subsection{Datasets}
The experiments are conducted on the publicly available DIBCO datasets \cite{pratikakis2011icdar}. We train our model on DIBCO 2009 \cite{gatos2009icdar}, DIBCO 2013 \cite{pratikakis2013icdar}, H-DIBCO 2012 \cite{pratikakis2012icfhr} and H-DIBCO 2014\cite{ntirogiannis2014icfhr2014} datasets. On the other hand, Challenging historical dataset like H-DIBCO 2016 \cite{pratikakis2016icfhr2016}  and DIBCO 2011 \cite{shahab2011icdar}  are selected for evaluation purposes. we resizes the images from these datasets to patches of size $256 \times 256$ before feeding to our model. To evaluate the performance of our methods, we adopt four evaluation metrics. They are F-measure, pseudo F-measure ($F_{ps}$), distance reciprocal distortion metric (DRD), and the peak signal-to-noise ratio (PSNR). Similar to \cite{bhunia2019improving}, we augment the training patches by rotating with an angle of 90, 180 and 270 degrees.

\subsection{Implementation Details}
We have implemented the entire model in Pytorch \cite{paszke2017automatic} and the experiments were done on a server having Nvidia Titan X GPU with 12 GB of memory. We have adapted step-wise training protocol for training our model. At first, we train ATANet for 15 epochs and generates the noisy version of the clean images. After that, we freeze the network such that the weights of the network does not alter. Next, we train UDBNet on generated noisy images for 20 epochs to generate its corresponding binary clean image. Then, We unfreeze the ATANet network. In next stage, we jointly train both the networks along with the joint discriminator for around $10$ epochs in the alternative manner. At last, we fine-tune the model for 30 epochs. During the couple training, ATANet tries to generate more challenging adversarial samples that are used as a pseudo image pair for training the UDBNet. Thus, the training procedure helps the model to learn various degradation including aging effects, noises etc. During training, we use Adam optimizer with the learning rate of $0.0001$. We take $\lambda_S$ = 0.5, $\lambda_C$ =10 and $\lambda_{L2}$ = 100 throughout the experiment.

 \begin{table*}[ht]    
     \centering
     \caption{Comparsion of Our method with Baseline Methods}
         \scalebox{1.5}{
         \begin{tabular}{l|c|c|c|c}
         \hline
         \textbf{Methods} & \textbf{F-Measure} & \pmb{$F_{PS}$} & \textbf{PSNR} & \textbf{DRD} \\
         \hline
         \textbf{UDBNet-CL } & 92.7 & 95.8 & 19.9 & 2.6 \\
         \hline
         \textbf{UDBNet-GRL} & 93.2 & 96.0 & 20.1 & 2.4 \\
         \hline
         \textbf{Ours} & \textbf{93.4} & \textbf{96.2} & \textbf{20.1} & \textbf{2.2} \\
         \hline
         \end{tabular}}
     \label{tab:My-img1}
 \end{table*}

\subsection{Baselines Methods}
In this section, We present two alternative baselines to justify the effectiveness of our methods :

\textbf{UDBNet-CL :} The joint discriminator exploits domain confusion loss \cite{ganin2016domain} to address the limitation described in the section 
\ref{section:Limitations of Base Model}. The domain confusion loss gives equal importance to ATANet and UDBNet.

\textbf{UDBNet-GRL :} The Gradient Reversal layer \cite{ganin2015unsupervised} ensures that the adversarial discriminator views the two domains identically. Here, the joint discriminator utilize the Gradient Reversal layer.

\begin{figure}[]
    
	\begin{center}
		\includegraphics[width=1\linewidth]{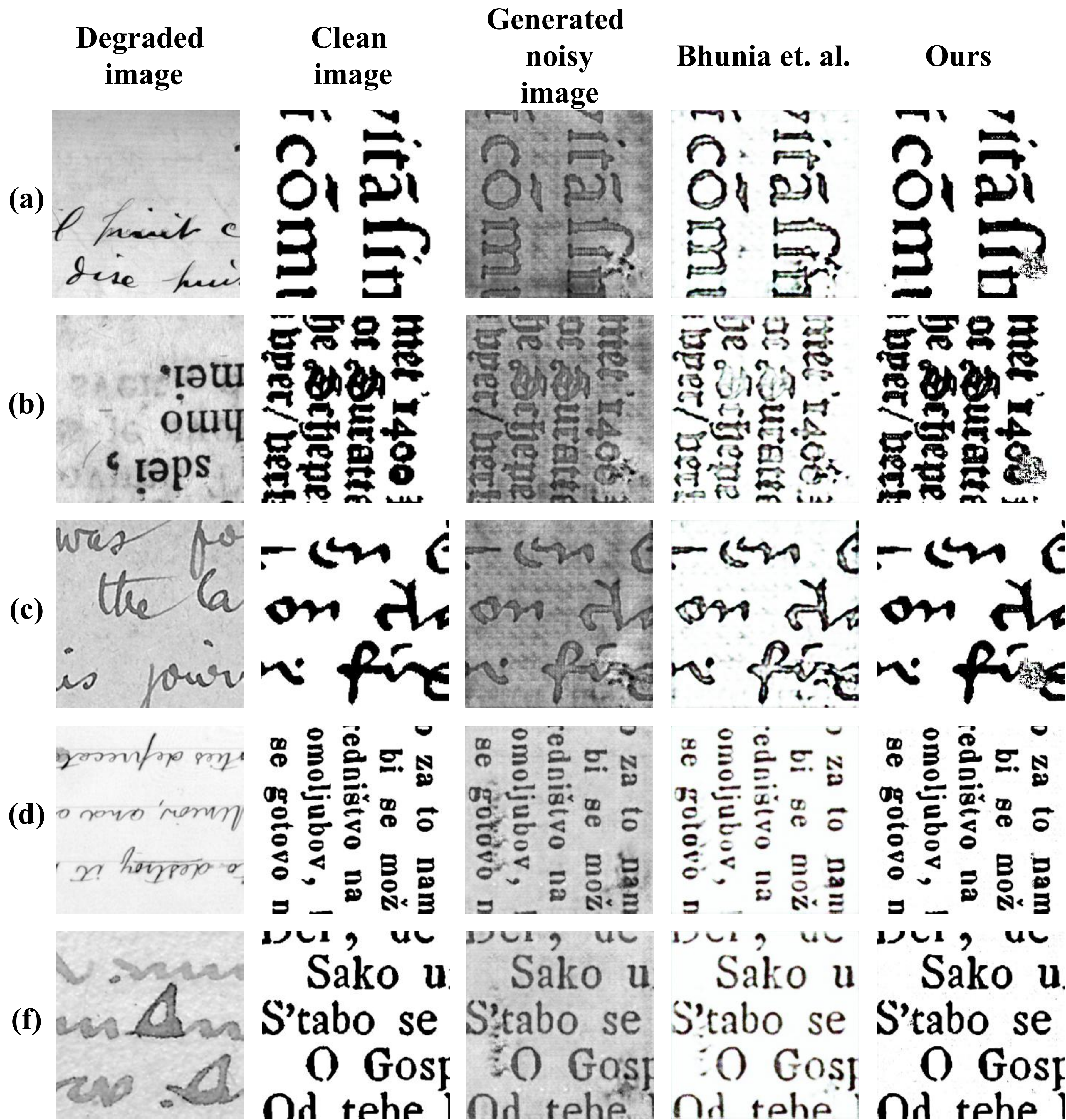}
	\end{center}
     \caption{Comparison of the qualitative results of predicted binarized images by Bhunia \etal \cite{bhunia2019improving} and our framework on the evaluation set}
	\label{fig:img:}
\end{figure}

\subsection{Performance Analysis}
\begin{figure}[]
    \begin{center}
		\includegraphics[width=0.9\linewidth]{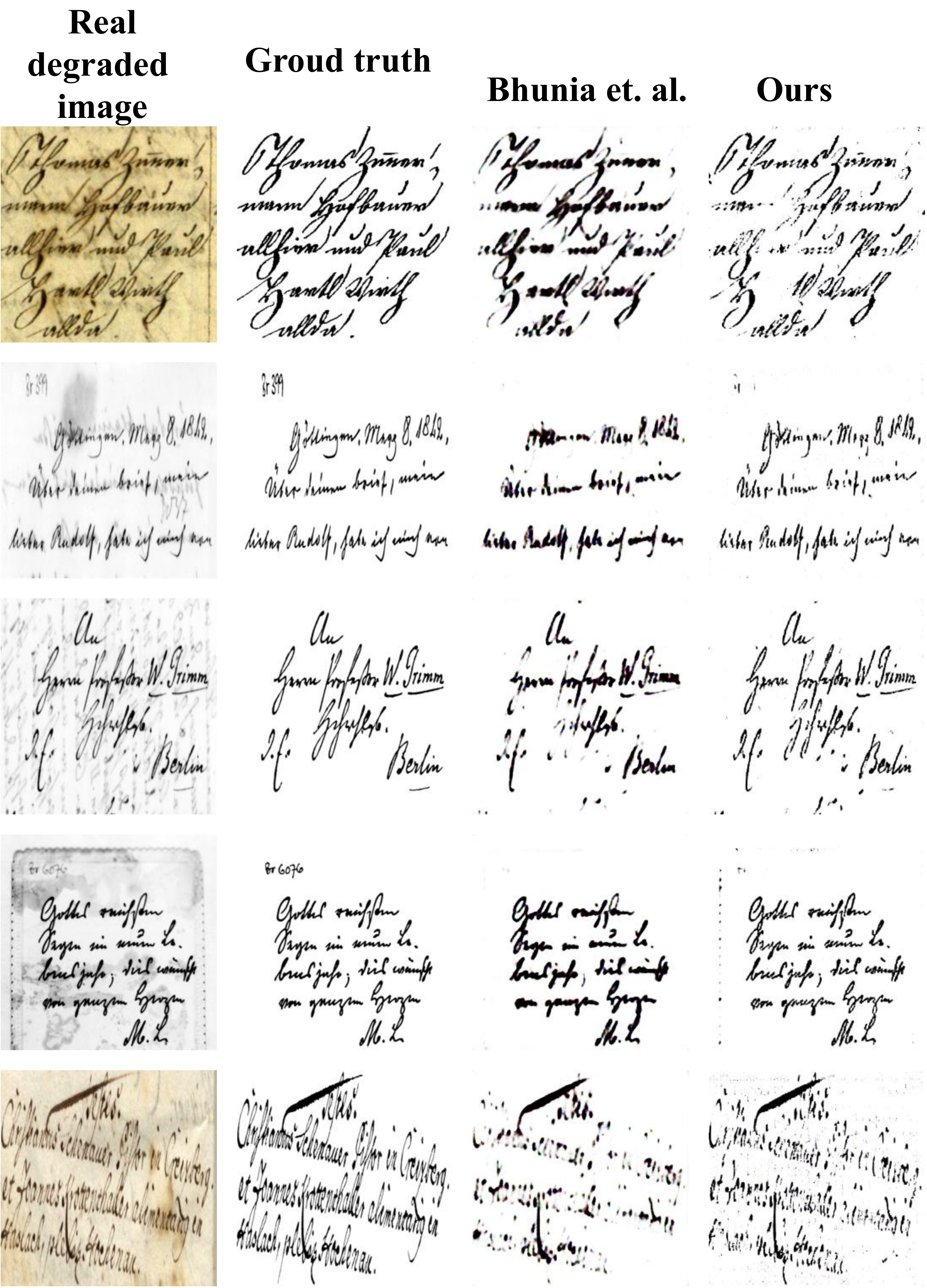}
	\end{center}
	\caption{ Binarization results on real test images by passing through UDBNet.}
	\label{fig:img:1}
\end{figure}

From Table \ref{tab:My-img1}, we observe that UDBNet-CL has achieved an improvement of $1.3\%$ and $0.4\%$ in F-Measure from  DeepOtsu \cite{he2019deepotsu} and Bhunia \cite{bhunia2019improving} on  H-DIBCO 2016 \cite{pratikakis2016icfhr2016} dataset. On the other hand, UDBNet-GRL shows better performance than UDBNet-CL, an improvement of $1.8\%$ and $0.9\%$ in F-Measure from  DeepOtsu \cite{he2019deepotsu} and Bhunia \cite{bhunia2019improving}  H-DIBCO 2016 \cite{pratikakis2016icfhr2016}. Our method outperformed all the previous state-of-the-art methods because  DeepOtsu \cite{he2019deepotsu} just uses stack refinement blocks and Bhunia \cite{bhunia2019improving} simply generates synthetic uncontrolled noisy image samples for training to improve the performance. In contrast, Our ATANet generates realistic degraded images including hard samples and also guides UDBNet to adopt to real noise distribution as depicted in Figures \ref{fig:img:} and \ref{fig:img:1}. However, out of three proposed approaches including baselines, the flipped label approach (ours) is found to be the best in all the four evaluation criteria because of its learning strategy. From Table \ref{tab:My-img}, it is obvious that our method has achieved significant improvement of $2.0\%$, $1.9\%$, $0.5\%$ from DeepOtsu \cite{he2019deepotsu} and $1.1\%$, $0.6\%$, $0.2\%$ from Bhunia \cite{bhunia2019improving} in F-Measure, $F_{PS}$ and PSNR criteria on  H-DIBCO 2016 \cite{pratikakis2016icfhr2016} dataset. Also, our method has shown improved performance of $1.9\%$, $1.5\%$ and $0.3\%$ in F-Measure, $F_{PS}$, PSNR than Vo \cite{vo2018binarization} on DIBCO 2011 dataset. In both the cases, the low DRD value of our method implies the robustness regarding visual distortion.

 \begin{table*}[ht]
     \centering
     \caption{Quantative results on H-DIBCO 2016 and DIBCO 2011 dataset}
     \scalebox{1.2}{
     \begin{tabular}{l|c|c|c|c|c|c|c|c}
     \hline
     & \multicolumn{4}{c|}{} & \multicolumn{4}{c}{}\\[-0.8em]
     \multirow{2}{*}{\textbf{Methods}} & \multicolumn{4}{c|}{\textbf{H-DIBCO 2016 Dataset}} & \multicolumn{4}{c}{\textbf{DIBCO 2011 Dataset}} \\[0.2em] \cline{2-9}
     & & & & & & & &\\[-0.9em]
     & \textbf{F-Measure} & \pmb{$F_{PS}$} & \textbf{PSNR} & \textbf{DRD} & \textbf{F-Measure} & \pmb{$F_{PS}$} & \textbf{PSNR} & \textbf{DRD} \\
     \hline
     Otsu \cite{otsu1979threshold} & 86.6 & 89.9 & 17.8 & 5.6 & 82.1 & 84.8 & 15.7 & 9.0\\
     \hline
     Sauvola \cite{sauvola2000adaptive} & 84.6 & 88.4 & 17.1 & 6.3 & 82.1 & 87.7 & 15.6 & 8.5\\
     \hline
     Howe\cite{howe2013document} & 87.5 & 92.3 & 18.1 & 5.4 & 91.7 & 92.0 & 19.3 & 3.4\\
     \hline
     Su \cite{su2010binarization} & 84.8 & 88.9 & 17.6 & 5.6 & 87.8 & 90.0 & 17.6 & 4.8\\
     \hline
     Jia \cite{jia2018degraded}& 90.5 & 93.3 & 19.3 & 3.9 & 91.9 & 95.1 & 19.0 & 2.6\\
     \hline
     Vo \cite{vo2018robust}& 87.3 & 90.5 & 17.5 & 4.4 & 88.2 & 90.3 & 20.1 & 2.9\\
     \hline
     Vo \cite{vo2018binarization}& 90.1 & 93.6 & 19.0 & 3.5 & 93.3 & 96.4 & 20.1 & 2.0\\
     \hline
     Westphal \cite{westphal2018document}& 88.8 & 92.5 & 18.4 & 3.9 & - & - & - & -\\
     \hline
     DeepOtsu \cite{he2019deepotsu}& 91.4 & 94.3 & 19.6 & 2.9 & 93.4 & 95.8 & 19.9 & 1.9\\
     \hline
     Bhunia \cite{bhunia2019improving} & 92.3 & 95.4 & 19.9 & 2.7 & 93.7 & 96.8 & 20.1 & 1.8\\
     \hline
     \textbf{Ours} & \textbf{93.4} & \textbf{96.2} & \textbf{20.1} & \textbf{2.2} & \textbf{95.2} & \textbf{97.9} & \textbf{20.4} &\textbf{1.5}\\
     \hline
     \end{tabular}}
     \label{tab:My-img}
 \end{table*}

\section{Conclusion} \label{conclusion}
In this paper, we have proposed a novel approach towards document
binarization by introducing three-player min-max adversarial
game. We introduce a joint discriminator which tries to couple
the Adversarial Texture Augmentation Network
(ATANet) and Unsupervised Document Binarization Network
(UDBNet) so that it can tackle the dataset bias problem and perform well on the real degraded document image. The proposed framework is simple and easy to implement. We demonstrate the effectiveness of our system by conducting experiments on publicly available DIBCO datasets. The results of the experiment show the superiority of our proposed model over the existing methods.


\bibliographystyle{IEEEtran}
\bibliography{IEEEexample}
%



\end{document}